\documentclass[runningheads]{llncs}
\usepackage{graphicx}

\usepackage{amsmath}
\usepackage{amsfonts}
\usepackage{amssymb}

\usepackage{algorithm}
\usepackage{algorithmicx}
\usepackage{algpseudocode}
\usepackage{xcolor}
\usepackage{comment}

\usepackage{subcaption}

\usepackage[hidelinks]{hyperref}

\usepackage{booktabs}
\usepackage{rotating}

\begin{document}
\title{The DLCC Node Classification Benchmark\\for Analyzing Knowledge Graph Embeddings}

\titlerunning{The DLCC Node Classification Benchmark}
\author{Jan Portisch\inst{1,2}\orcidID{0000-0001-5420-0663} \and
Heiko Paulheim\inst{2}\orcidID{0000-0003-4386-8195}}
\authorrunning{Portisch \& Paulheim}

\institute{SAP SE, Walldorf, Germany\\
\email{jan.portisch@sap.com}\\
\and
Data and Web Science Group, University of Mannheim, Germany\\
\email{\{jan,heiko\}@informatik.uni-mannheim.de}}
\maketitle

\setcounter{footnote}{0}
\begin{abstract}
Knowledge graph embedding is a representation learning technique that projects entities and relations in a knowledge graph to continuous vector spaces.
Embeddings have gained a lot of uptake and have been heavily used in link prediction and other downstream prediction tasks. 
Most approaches are evaluated on a single task or a single group of tasks to determine their overall performance. The evaluation is then assessed in terms of how well the embedding approach performs on the task at hand. Still, it is hardly evaluated (and often not even deeply understood) what information the embedding approaches are \emph{actually} learning to represent.

To fill this gap, we present the DLCC (Description Logic Class Constructors) benchmark, a resource to analyze embedding approaches in terms of which kinds of classes they can represent. Two gold standards are presented, one based on the real-world knowledge graph DBpedia and one synthetic gold standard. In addition, an evaluation framework is provided that implements an experiment protocol so that researchers can directly use the gold standard. 
To demonstrate the use of DLCC, we compare multiple embedding approaches using the gold standards. We find that many DL constructors on DBpedia are actually learned by recognizing different correlated patterns than those defined in the gold standard and that specific DL constructors, such as cardinality constraints, are particularly hard to be learned for most embedding approaches.

\end{abstract}
\section{Introduction}
Knowledge graph embeddings are projections of entities and relations to continuous vector spaces. They have been proposed for various purposes and are typically evaluated on task-specific gold standards such as FB15k and WN18~\cite{DBLP:conf/nips/BordesUGWY13} for link prediction, kgbench for node classification \cite{DBLP:conf/esws/BloemWBB21}, or GEval \cite{DBLP:conf/esws/PellegrinoAGRC20,DBLP:conf/esws/PellegrinoCGR19} for machine learning tasks such as classification, regression, or clustering. The benchmarks frequently come with their own evaluation protocol.

\begin{figure}[t]
    \centering
    \begin{subfigure}[t]{0.49\textwidth}
        \includegraphics[height=5.5cm]{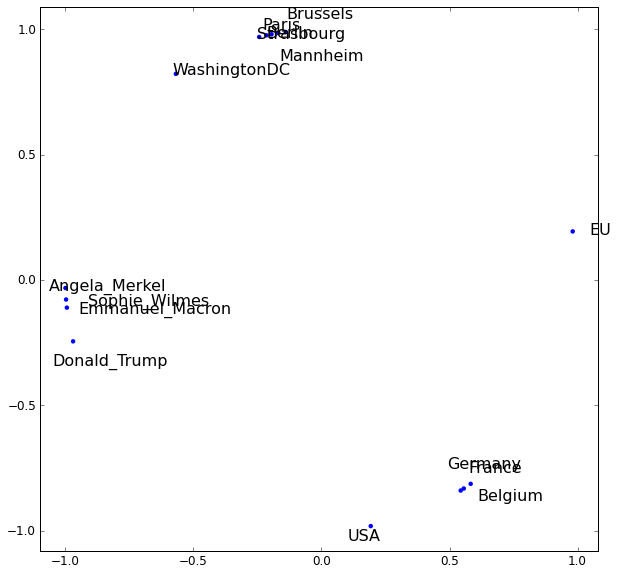}
        \caption{Good class separation}
    \end{subfigure}
    \begin{subfigure}[t]{0.49\textwidth}
        \includegraphics[height=5.5cm]{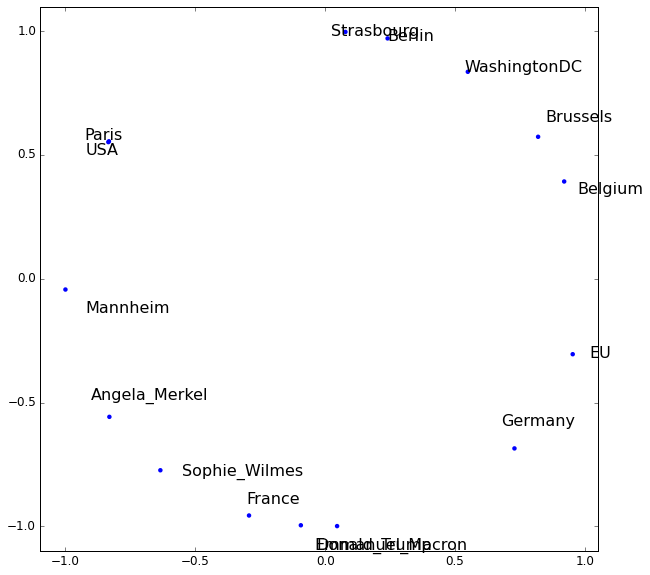}
        \caption{Bad class separation}
    \end{subfigure}
    \caption{Two Example Embeddings. The left-hand side embedding shows a good class separation of persons, countries, and cities, whereas the right-hand side one does not.}
    \label{fig:example_embeddings}
\end{figure}

Independent of the original benchmark task, knowledge graph embeddings are generally versatile so that they can be used for multiple tasks~\cite{DBLP:journals/semweb/PortischHP22}. 
While the performance of embeddings in downstream tasks is often superior to other entity representation techniques, most, if not all, embedding approaches have in common that it is not ultimately clear \emph{what} is learned. For example, both for link prediction and for node classification, it is required that classes can be separated (e.g., persons, countries, and cities are clustered in the embedding space)~\cite{DBLP:journals/semweb/PortischHP22}, but so far, it has not been systematically evaluated which embedding methods can learn which kinds of class separations. Figure~\ref{fig:example_embeddings} shows an example of two embedding spaces with different qualities of class separation.

In this paper, we present the DLCC (for \emph{Description Logic Class Constructors}) dataset and an evaluation framework that help to better analyze and understand embedding approaches for specific DL constructors. There are four contributions of this paper: (1) A framework for the DLCC gold standard creation is presented, (2) two concrete gold standards are provided -- a real graph-based gold standard and one based on synthetic knowledge graphs, (3) an evaluation framework is provided to easily evaluate and compare the class separation capabilities of embeddings, and (4) a preliminary analysis for different state of the art embedding approaches is provided. 

\section{Related Work}
In the area of link prediction (or knowledge base completion), the two well-known evaluation datasets FB15k and WN18~\cite{DBLP:conf/nips/BordesUGWY13} are both based on real datasets: FB15k is based on the Freebase dataset, and WN18 is based on WordNet~\cite{miller1998wordnet}. They were presented in the context of link prediction: 
Given a triple in the form \emph{(\textbf{h}ead, \textbf{r}elation, \textbf{t}ail)}, two prediction tasks \emph{(head, relation, ?)} and \emph{(?, relation, tail)} are created. The evaluation is performed by calculating the mean rank/HITS@10 for a list of proposals. 
Since it has been remarked that those datasets contain too many simple inferences due to inverse relations, the more challenging variants FB15k-237~\cite{toutanova2015observed} and WN18RR~\cite{DBLP:conf/aaai/DettmersMS018} have been proposed. More recently, evaluation sets based on larger knowledge graphs, such as YAGO3-10~\cite{DBLP:conf/aaai/DettmersMS018} and DBpedia50k/DBpedia500k~\cite{DBLP:conf/aaai/ShiW18}, have been introduced.

Bloem et al.~\cite{DBLP:conf/esws/BloemWBB21} introduce \emph{kgbench}, a node classification benchmark for knowledge graphs, which, like DLCC, comes with datasets in different sizes and predefined train/test splits. Unlike DLCC, kgbench is based on real-world datasets. Therefore, it is suitable to evaluate and compare the quality of different embedding approaches on real-world tasks but does not provide any insights into what these embedding approaches are capable of representing.

Alshagari et al.~\cite{DBLP:conf/aaaiss/AlshargiSSS19} present a framework for ontological concepts covering three aspects: (i) categorization, (ii) hierarchy, and (iii) logic validation. The framework can be used for language models and for knowledge graph embeddings. The work presented in this paper differs in that it goes beyond explicit DBpedia types. The evaluation of this paper is, therefore, of analytical rather than descriptive nature. Moreover, the task sets of DLCC are significantly larger and more comprehensive.

Ristoski et al.~\cite{DBLP:conf/semweb/RistoskiVP16} provide a collection of benchmarking datasets for machine learning, including classification, clustering, and regression tasks. Later, the GEval framework~\cite{DBLP:conf/esws/PellegrinoAGRC20,DBLP:conf/esws/PellegrinoCGR19} was introduced to provide a standardized evaluation protocol for this dataset. The evaluation datasets are based on DBpedia. Internally, the embeddings are processed by different downstream classification, regression, or clustering algorithms. The evaluation framework presented in this paper is similar to GEval in that it also evaluates multiple classifiers given a concept vector input.

Melo and Paulheim~\cite{DBLP:conf/esws/MeloP17} provide a method for synthesizing benchmark datasets for link and entity type prediction, which are used in conjunction with a fixed ontology. Their goal is to mimic the characteristic of existing knowledge graphs in terms of distributions and patterns.

\section{Covered DL Constructors}
The aim of this paper is to provide a benchmark for analyzing which kinds of constructs in a knowledge graph can be recognized by different embedding methods. To that end, we define class labels using different DL constructors. Later on, we apply classification algorithms to analyze how well the differently labeled classes can be separated using different embedding algorithms.

\paragraph{Ingoing and Outgoing Relations}
All entities that have a particular outgoing or ingoing relations (e.g., \emph{everything that has a location}). 
\begin{equation}
    \exists r.\top
    \label{eq:relation_out}
\end{equation}
\begin{equation}
    \exists r^{-1}.\top
    \label{eq:relation_in}
\end{equation}
\begin{equation}
     \exists r.\top \sqcup \exists r^{-1}.\top
    \label{eq:relation_inout}
\end{equation}
where $r$ is bound to a particular relation.\footnote{We use $r$ to denote a particular relation, whereas $R$ denotes \emph{any} relation.}

\paragraph{Relations to Particular Individuals} 
All entities that have a relation (in any direction) to a particular individual (e.g., \emph{everything that is related to New York City}).
\begin{equation}
    \exists R.\left\{e\right\} \sqcup \exists R^{-1}.\left\{e\right\}
    \label{eq:relation_individual}
\end{equation}
where $R$ is \emph{not} bound to a particular relation.
Those relations can also span two (or more\footnote{For reasons of scalability, we restrict the provided gold standard to two hops.} hops):
\begin{equation}
    \exists R_1.(\exists R_2.\left\{e\right\}) \sqcup \exists R_1^{-1}.(\exists R_2^{-1}\left\{e\right\})
    \label{eq:relation_individual_2hop}
\end{equation}

\paragraph{Particular Relations to Particular Individuals}
All entities that have a particular relation to a particular individual (e.g., \emph{movies directed by Steven Spielberg}). 
\begin{equation}
    \exists r.\left\{e\right\}
    \label{eq:particular_relation_individual1}
\end{equation}

\paragraph{Qualified Restrictions}
All entities that have a particular relation to an individual of a given type (e.g., \emph{all people married to soccer players}).
\begin{equation}
    \exists r.T
    \label{eq:qualified_restriction_out}    
\end{equation}
\begin{equation}
    \exists r^{-1}.T
    \label{eq:qualified_restriction_in}    
\end{equation}
If types are modeled as a normal relation in the graph (i.e., \texttt{rdf:type} is yet another relation), we can reformulate Equation~\ref{eq:qualified_restriction_out} and~\ref{eq:qualified_restriction_in} to
\begin{equation}\tag{7a}
    \exists r.(\exists \texttt{rdf:type}.T)
\end{equation}
\begin{equation}\tag{8a}
    \exists r^{-1}.(\exists \texttt{rdf:type}.T)
\end{equation}
In that case, it behaves equally to a chained variant of Equation~\ref{eq:particular_relation_individual1}.

\paragraph{Cardinality Restrictions of Relations}
All entities that have at least or at most $n$ relations of a particular kind (e.g., \emph{people who have at least two citizenships}). Here, we depict only the \emph{lower bound} variant because the corresponding decision problem is between the two variants (entities that fall below the bound, i.e., adhere to the upper bound, are in the negative example set).\footnote{The fact that most KGs follow the open-world assumption is neglected here since we test for the presence/absence of patterns.}
\begin{equation}
    \geq 2 r.\top
    \label{eq:min_cardinality_out}    
\end{equation}
\begin{equation}
    \geq 2 r^{-1}.\top
    \label{eq:min_cardinality_in}    
\end{equation}

\paragraph{Qualified Cardinality Restrictions}
Qualified cardinality restrictions combine qualified restrictions with cardinalities (e.g., people who have published at least two science fiction novels).
\begin{equation}
    \geq 2 r.T
    \label{eq:qualified_cardinality_out}    
\end{equation}
\begin{equation}
    \geq 2 r^{-1}.T
    \label{eq:qualified_cardinality_in}    
\end{equation}

\noindent Table~\ref{tab:tc-overview} summarizes the DL constructors for which test cases were built. 

\section{Approach}
For the twelve test cases in Table~\ref{tab:tc-overview}, we create positive examples (i.e., those which fall into the respective class) and those which do not (under closed-world semantics). For example, for tc01, we would generate a set of positive instances for which $\exists r.\top$ holds and a set of negative instances for which $\nexists r.\top$ holds. We then evaluate how well these two classes can be separated, given the embedding vectors of the positive and negative instances. For that, we split the examples into a training and testing partition, we train binary classifiers on the training subset of the examples, and evaluate their performance on the test subset. 

\begin{figure}[t]
    \centering
    \includegraphics[width=0.85\textwidth]{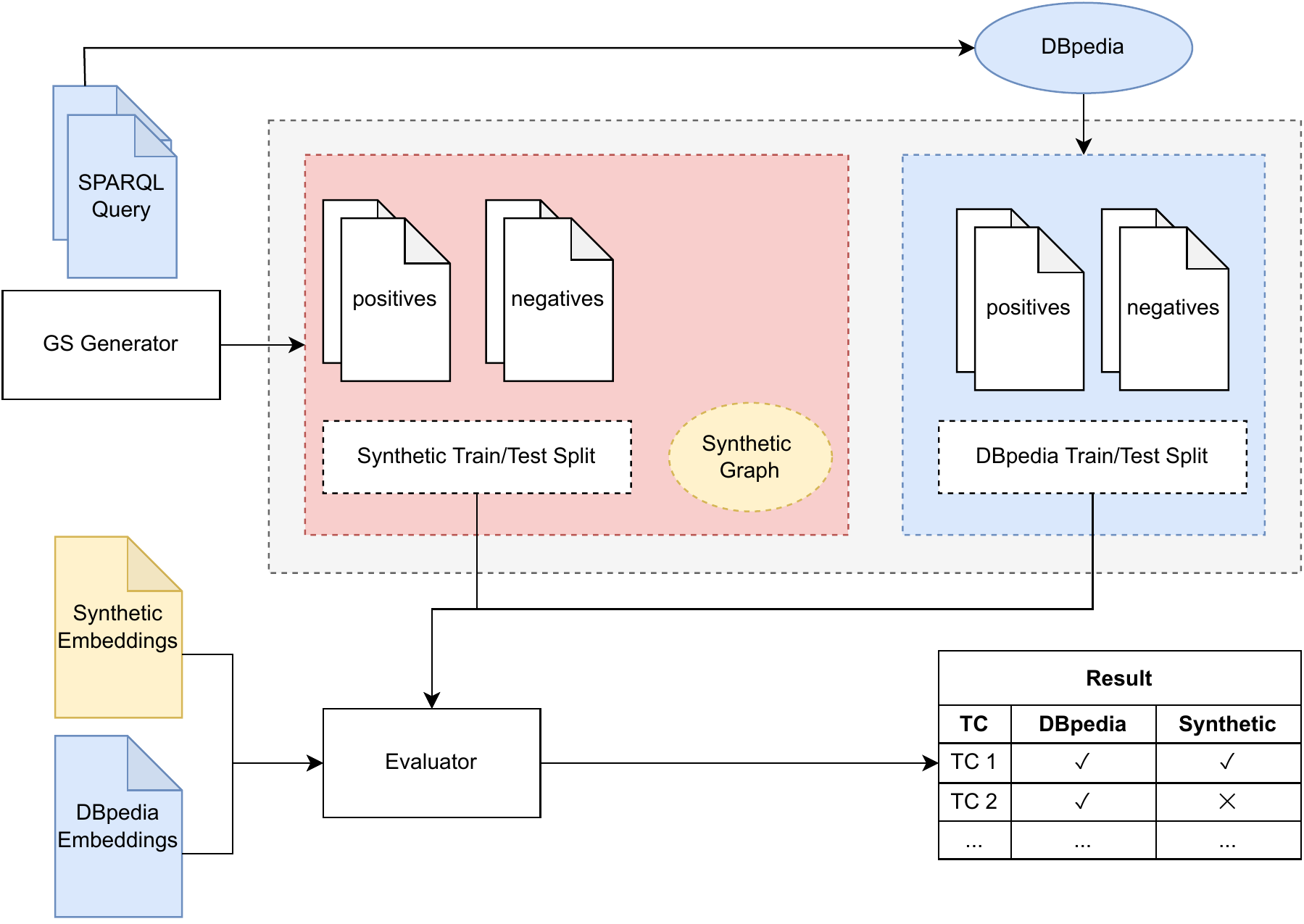}
    \caption{Overview of the Approach}
    \label{fig:gs-approach}
\end{figure}

The approach is visualized in Figure~\ref{fig:gs-approach}: A gold standard generator generates a set of positive and negative URIs, as well as a fixed train/test split. The approach presented in this paper allows to generate custom gold standards -- however, a contribution of this paper is also to provide a pre-calculated gold standard. This pre-calculated gold standard can be used to guarantee reproducibility. Officially published gold standards are versioned to allow for future improvements. In this paper, we present version \texttt{v1} of the gold standard.

\begin{table}[t]
\centering
\caption{Overview of the Test Cases}
\label{tab:tc-overview}
\begin{tabular}{l|l|}
         Test Case & DL Expression  \\
         \hline
         \hline
         tc01 & $\exists r.\top$  \\
         tc02 & $\exists r^{-1}.\top$  \\
         tc03 & $\exists r.\top \sqcup \exists r^{-1}.\top$ \\
         \hline
         tc04 & $\exists R.\left\{e\right\} \sqcup \exists R^{-1}.\left\{e\right\}$  \\
         tc05 & $\exists R_1.(\exists R_2.\left\{e\right\}) \sqcup \exists R_1^{-1}.(\exists R_2^{-1}\left\{e\right\})$  \\
         \hline
         tc06 & $\exists r.\left\{e\right\}$ \\
         \hline
         tc07 & $\exists r.T$ \\
         tc08 & $\exists r^{-1}.T$ \\
         \hline
         tc09 & $\geq 2 r.\top$ \\
         tc10 & $\geq 2 r^{-1}.\top$ \\
         \hline
         tc11 & $\geq 2 r.T$ \\
         tc12 & $\geq 2 r^{-1}.T$ \\
         \end{tabular}
\end{table}

A user provides embeddings in a simple textual format and provides them together with the training data as input to the evaluator. 
The evaluator trains multiple classifiers and evaluates them on the selected gold standard using the provided vectors as classification input. 
The program then calculates multiple statistics in the form of CSV files that can be further analyzed in a spreadsheet program or through data analysis frameworks such as pandas\footnote{\url{https://pandas.pydata.org/}}. These analyses help the user to understand how well the provided vectors are performing on a particular DL constructor.

\subsection{Gold Standard Generator}
The gold standard generator is publicly available\footnote{\url{https://github.com/janothan/DL-TC-Generator}}. It is implemented as a Java maven project. The generator can generate either a DBpedia benchmark (see Subsection~\ref{ssec:dbpedia-benchmark}) or a synthetic one (see Subsection~\ref{ssec:synthetic-benchmark}). Any DBpedia version can be used; the user merely needs to provide a SPARQL endpoint. A comprehensive set of unit tests ensures a high code quality. The generator automatically generates a fixed train-test split for the evaluation framework or any other downstream application. The split is configurable; for the pre-generated gold standards, an 80-20 split is used. The resulting gold standard is balanced -- i.e., the number of positives equals the number of negatives -- and the train and test partitions are stratified. Hence, any classifier which achieves an accuracy significantly above 50\% is capable of learning the test case's problem type from the vectors to some extent.

It is important to note that the generator only needs to be run by users who want to build their own gold standards. The typical user would merely download\footnote{DOI: \texttt{10.5281/zenodo.6509715}; GitHub link for the latest version. \url{https://github.com/janothan/DL-TC-Generator/tree/master/results}} the official gold standard files online. We recommend using the pre-calculated gold standards to ensure comparability across publications. 

\subsection{Evaluation Framework}
The evaluator is publicly available\footnote{\url{https://github.com/janothan/dl-evaluation-framework}} as well together with usage examples. It is implemented in Python and can be easily used in a Jupyter notebook. A comprehensive set of unit tests ensures a high code quality.

The standard user can directly download the gold standard and use the evaluation framework. To test class separability, the evaluation framework currently runs six machine learning classifiers:\footnote{The evaluation framework is not restricted to the set of classifiers listed here. New classifiers can be easily added if desired.} (1) decision trees, (2) na\"{i}ve Bayes, (3) KNN, (4) SVM, (5) random forest, and (6) a multilayer perceptron network. The framework uses the default configurations of the sklearn library\footnote{\url{https://scikit-learn.org/stable/index.html}}.

After training and evaluation, the framework persists multiple CSV files per test case as well as higher-level aggregate CSV files. Examples of such CSV files are a file listing the accuracy per classifier and per test case or a file listing the accuracy of the best classifier per test case. In the case of DBpedia, test cases are created for multiple domains, and the results can be analyzed on the level of each domain separately or in an aggregated manner on the level of the test case. 

\section{Benchmarks}
We currently provide two benchmarks, while the framework described above allows for generating customized benchmarks.

\subsection{DBpedia Benchmark}
\label{ssec:dbpedia-benchmark}
We use the DBpedia knowledge graph to create test cases.\footnote{We used DBpedia version 2021-09. The generator can be configured to use any DBpedia SPARQL endpoint if desired.} We created SPARQL queries for each test case (see Table~\ref{tab:tc-overview}) to generate positives, negatives, and hard negatives. The latter are meant to be less easily distinguishable from the positives and are created by variations such as softening the constraints in the class constructor or switching subject and object in the constraint. For example, for qualified relations, a positive example would be a person playing in a team which is a basketball team. A simple negative example would be any person not playing in a basketball team, whereas a hard negative example would be any person playing in a team that is not a basketball team.

Query examples for every test case in the people domain are provided in Table~\ref{tab:sparql-tcs-1}. The framework uses slightly more involved queries to vary the size of the result set and to better randomize results.

\begin{table}
    \centering
    \scriptsize
    \begin{tabular}{l|p{3.5cm}|p{3.6cm}|p{3.8cm}}
         \textbf{TC} & \textbf{Query Positive} & \textbf{Query Negative} & \textbf{Query Negative\newline(hard)}\\
         \hline
         tc01 & 
         \texttt{SELECT DISTINCT(?x) WHERE \{\newline \phantom{XX}?x a dbo:Person . \newline \phantom{XX}?x dbo:child ?y . \} } & 
         \texttt{SELECT DISTINCT(?x) WHERE \{\newline \phantom{XX}?x a dbo:Person .\newline \phantom{XX}FILTER(NOT EXISTS \{\newline \phantom{XXXX}?x dbo:child ?z\})\} } &
         \texttt{SELECT DISTINCT(?x) WHERE \{\newline \phantom{XX}?x a dbo:Person .\newline \phantom{XX}?y dbo:child ?x. \newline \phantom{XX}FILTER(NOT EXISTS \{\newline \phantom{XXXX}?x dbo:child ?z\})\} }
         \\
         \hline
         tc02 & 
         \multicolumn{3}{c}{Analogous to tc01 (inverse case).}
         \\
         \hline
         tc03 & 
         \texttt{SELECT DISTINCT(?x) WHERE \{\newline\{ ?x a dbo:Person . \newline\phantom{XX}?x dbo:child ?y\} UNION \newline\{ ?x a dbo:Person . \newline\phantom{XX}?y dbo:child ?x\}\} } & 
         \texttt{SELECT COUNT(?x) WHERE \{\newline \phantom{XX}?x a dbo:Person . \newline\phantom{XX}FILTER(NOT EXISTS\{\newline\phantom{XXXX}?x dbo:child ?y\}\newline\phantom{XX}AND NOT EXISTS \{\newline\phantom{XXXX}?z dbo:child ?x\})\}}
         & --\\
         \hline
         tc04 & 
         \texttt{SELECT DISTINCT(?x) WHERE \{\newline\{ ?x a dbo:Person . \newline\phantom{XX}?x ?y dbr:New\_York\_City\} UNION \newline\{ ?x a dbo:Person . \newline\phantom{XX}dbr:New\_York\_City ?y ?x\}\}} 
         &
         \texttt{SELECT DISTINCT(?x) WHERE \{\newline \phantom{XX}?x a dbo:Person . \newline\phantom{XX}FILTER(NOT EXISTS\{\newline\phantom{XXXX}?x ?y dbr:New\_York\_City\}\newline\phantom{XX}AND NOT EXISTS \{\newline\phantom{XXXX}dbr:New\_York\_City ?y ?x\})\}}
         &
         \texttt{SELECT DISTINCT(?x) WHERE \{\{\newline
         \phantom{XX}?x a dbo:Person .\newline
         \phantom{XX}?x ?y1 ?z .\newline
         \phantom{XX}?z ?y2 dbr:New\_York\_City \}\newline
         \phantom{XX}UNION \{ \newline
         \phantom{XX}?x a dbo:Person .\newline
         \phantom{XX}?z ?y1 ?x .\newline
         \phantom{XX}dbr:New\_York\_City ?y2 ?z \}\newline
         \phantom{XX}FILTER(NOT EXISTS\newline
         \phantom{XXXX}\{?x ?r dbr:New\_York\_City\} \newline
         \phantom{XX}AND NOT EXISTS\newline
         \phantom{XXXX}\{dbr:New\_York\_City ?s ?x\})\}}
         \\
         \hline
         tc05 &
         \multicolumn{3}{c}{Analogous to tc04 (inverse case).}
         \\
         \hline
         tc06 & \texttt{SELECT DISTINCT(?x) WHERE \{\newline
         \phantom{XX}?x a dbo:Person .\newline
         \phantom{XX}?x dbo:birthPlace \newline\phantom{XXXX}dbr:New\_York\_City \}}
         &
         \texttt{SELECT DISTINCT(?x) WHERE \{\newline
         \phantom{XX}?x a dbo:Person .\newline
         \phantom{XX}FILTER(NOT EXISTS\{\newline
         \phantom{XXXX}?x dbo:birthPlace\newline\phantom{XXXXXX}dbr:New\_York\_City \})\}}
         &
         \texttt{SELECT DISTINCT(?x) ?r WHERE \{\{\newline
         \phantom{XXXX}?x a dbo:Person .\newline
         \phantom{XXXX}?x dbo:birthPlace ?y .\newline
         \phantom{XXXX}dbr:New\_York\_City ?r ?x .\newline
         \phantom{XXXX}FILTER(?y!=dbr:New\_York\_City)\}\newline
         \phantom{XX}UNION \{\newline
         \phantom{XXXX}?x a dbo:Person .\newline
         \phantom{XXXX}?x dbo:birthPlace ?y .\newline
         \phantom{XXXX}?x ?r dbr:New\_York\_City .\newline
         \phantom{XXXX}FILTER(?y!=dbr:New\_York\_City)\}\}}\\
%
%
         \hline
         tc07 &
         \texttt{SELECT DISTINCT(?x) WHERE \{\newline
         \phantom{XX}?x a dbo:Person . \newline
         \phantom{XX}?x dbo:team ?y . \newline
         \phantom{XX}?y a dbo:BasketballTeam \}}
         &
         \texttt{SELECT DISTINCT(?x) WHERE \{\newline
         \phantom{XX}?x a dbo:Person . \newline
         \phantom{XX}FILTER(NOT EXISTS\{\newline
         \phantom{XXXX}?x dbo:team ?y . \newline
         \phantom{XXXX}?y a dbo:BasketballTeam\})\}}
         &
         \texttt{SELECT DISTINCT(?x) WHERE \{\newline
         \phantom{XX}?x a dbo:Person . \newline
         \phantom{XX}?x dbo:team ?z1 .\newline
         \phantom{XX}?x ?r ?z2 .\newline
         \phantom{XX}?z2 a dbo:BaseballTeam \newline
         \phantom{XX}FILTER(NOT EXISTS\{\newline
         \phantom{XXXX}?x dbo:team ?y . \newline
         \phantom{XXXX}?y a dbo:BasketballTeam \})\}}\\
         \hline
         tc08 &
         \multicolumn{3}{c}{Analogous to tc07 (inverse case).}
         \\
    \hline
    tc09 & 
    \texttt{SELECT DISTINCT(?x) WHERE \{ \newline
    \phantom{XX}?x a dbo:Person .\phantom{XX}\newline
    \phantom{XX}?x dbo:award ?y1. \newline
    \phantom{XX}?x dbo:award ?y2.\newline
    \phantom{XX}FILTER(?y1!=?y2)\}}
    &
    \texttt{SELECT DISTINCT(?x) WHERE \{ \newline
    \phantom{XX}?x a dbo:Person .\phantom{XX}\newline
    \phantom{XX}FILTER(NOT EXISTS\{\newline
    \phantom{XXXX}?x dbo:award ?y1. \newline
    \phantom{XXXX}?x dbo:award ?y2.\newline
    \phantom{XXXX}FILTER(?y1!=?y2)\})\}}
    &
    \texttt{SELECT DISTINCT(?x) WHERE \{ \newline
    \phantom{XX}?x a dbo:Person .\phantom{XX}\newline
    \phantom{XX}?x dbo:award ?y .\newline
    \phantom{XX}FILTER(NOT EXISTS\{\newline
    \phantom{XXXX}?x dbo:award ?z. \newline
    \phantom{XXXX}FILTER(?y!=?z)\})\}}\\
    \hline
    tc10 &
     \multicolumn{3}{c}{Analogous to tc09 (inverse case).}
    \\
    \hline
    tc11 &
    \texttt{SELECT DISTINCT(?x) WHERE \{\newline
    \phantom{XX}?x a dbo:Person . \newline
    \phantom{XX}?x dbo:recordLabel ?y1 .\newline
    \phantom{XX}?y1 a dbo:RecordLabel .\newline
    \phantom{XX}?x dbo:recordLabel ?y2 .\newline
    \phantom{XX}?y2 a dbo:RecordLabel .\newline
    \phantom{XX}FILTER(?y1!=?y2)\}}
    &
    \texttt{SELECT DISTINCT(?x) WHERE \{\newline
    \phantom{XX}?x a dbo:Person . \newline
    \phantom{XX}FILTER(NOT EXISTS\{\newline
    \phantom{XXXX}?x dbo:recordLabel ?y1 .\newline
    \phantom{XXXX}?y1 a dbo:RecordLabel .\newline
    \phantom{XXXX}?x dbo:recordLabel ?y2 .\newline
    \phantom{XXXX}?y2 a dbo:RecordLabel .\newline
    \phantom{XXXX}FILTER(?y1!=?y2)\})\}}
    &    
    \texttt{SELECT DISTINCT(?x) WHERE \{\newline
    \phantom{XX}?x a dbo:Person . \newline
    \phantom{XX}?x dbo:recordLabel ?y1 .\newline
    \phantom{XX}?y1 a dbo:RecordLabel .\newline
    \phantom{XX}FILTER(NOT EXISTS\{\newline
    \phantom{XXXX}?x dbo:recordLabel ?y2 .\newline
    \phantom{XXXX}?y2 a dbo:RecordLabel .\newline
    \phantom{XXXX}FILTER(?y1!=?y2)\})\}}\\
    \hline
    tc12 &
     \multicolumn{3}{c}{Analogous to tc11 (inverse case).}
    \end{tabular}
    \caption{Exemplary SPARQL Queries for Class \texttt{Person}}
    \label{tab:sparql-tcs-1}
\end{table}

In total, we used six different domains: people (P), books (B), cities (C), music albums (A), movies (M), and species (S). This setup yields more than 200 hand-written SPARQL queries, which are used to obtain positives, negatives, and hard negatives; they are available online\footnote{\url{https://github.com/janothan/DL-TC-Generator/tree/master/src/main/resources/queries}} and can be easily extended, e.g., to add an additional domain. For each test case, we created differently sized (50, 500, 5000) balanced test sets.\footnote{The desired size classes can be configured in the framework.}

\subsection{Synthetic Benchmark}
\label{ssec:synthetic-benchmark}
The previous benchmark is realistic and well suited to compare approaches on differently typed DL constructors. 

However, the following aspects have to be considered: 
(1) DBpedia is a large knowledge graph; not every embedding approach can be used to learn an embedding for it (or not every researcher has the computational means to do so, respectively). 
(2) Depending on the DL constructor and the domain, not enough examples can be found on DBpedia.
(3) It cannot be precluded that patterns correlate; therefore, the fact that an embedding approach can learn a particular class can only be an indicator that it \emph{might} learn the underlying constructor pattern, but the results are not conclusive. Correlating properties, type biases for entities, etc., may lead to surprising results in some domains (see Section~\ref{subsec:dbpedia_vs_synthetic}). 

Therefore, we complement the DBpedia-based gold standard with a synthetic benchmark. The idea is to generate a graph that contains the DL constructors (positive and negative) of interest. The graph can be constructed to resemble the DBpedia graph statistically but can be significantly smaller (and contain a sufficient number of positives and negatives), and, by construction, side effects and correlations which exist in DBpedia can be mitigated to a large extent. 

The configurable parameters are \texttt{numClasses}, \texttt{numProperties}, \texttt{numInstances}, \texttt{branchingFactor}, \texttt{maxTriplesPerNode}, and \texttt{numNodesInterest} (all parameters are integers). 
The overall process is depicted in Algorithm~\ref{alg:ontology-creation}: First, a class tree with \texttt{numClasses} classes is constructed in a way that each class has at most \texttt{branchingFactor} children. 
Then, \texttt{numproperties} properties are generated. Each property is assigned to a range and domain from the class tree, whereby the first property has the root node as domain and range type so that every node can be involved in at least one triple statement. A skew can be introduced so that domain and range refer with a higher probability to a more general class than to a specific one. Lastly, we generate instances and assign them to a class as type, which is depicted in Algorithm~\ref{alg:ontology-creation}.

\begin{figure}[t]
    \centering
    \includegraphics[width=0.4\textwidth]{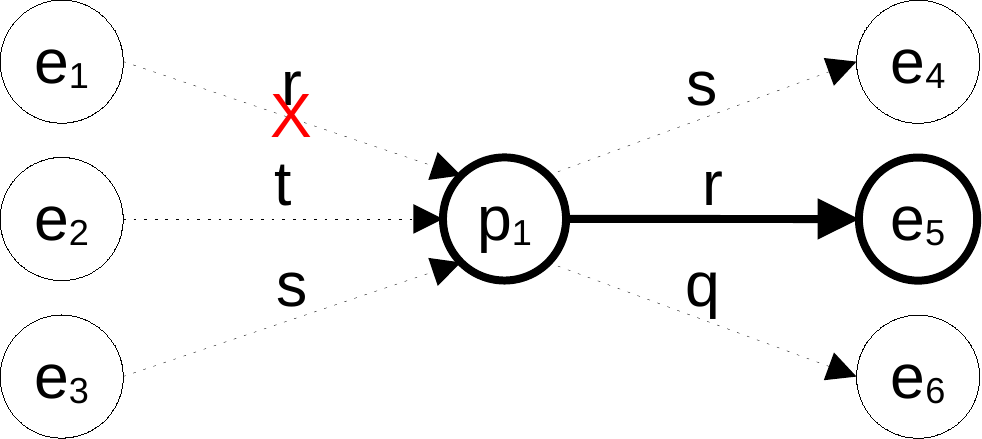}
    \caption{Illustration of the instance generation, using the class constructor $\exists r.T$. First, the pattern is instantiated for the positive example $p_1$ with the edge $(p_1,r,e_5)$. Then, random edges are inserted (dashed lines). The edge $(e_1,r,p_1)$ is removed, because it would turn $e_1$ into an additional positive example.}
    \label{fig:population}
\end{figure}
Once the ontology is created, \texttt{numNodesInterest} positives and negatives are generated (adhering to domain/range restrictions). Each class constructor is first initialized explicitly for the positive examples. Then, for each entity $e$ in the graph (i.e., positive and negative examples), $rand(n) \in [1, maxTriplesPerNode]$ random triples are generated, which have $e$ as a subject and adhere to the domain and range definitions, whereby it is checked that no additional positives are created, and no negatives are turned into positives accidentally (see Figure~\ref{fig:population}).

For version \texttt{v1} of the gold standard, \texttt{numClasses = 760}, \texttt{numProperties = 1355}, \texttt{numInstances = 10,000}, \texttt{branchingFactor = 5}, \texttt{maxTriplesPerNode = 11}, and \texttt{numNodesInterest = 1000} were chosen. The parameters were chosen to form graphs which are smaller than DBpedia but resemble the DBpedia graph statistically. Therefore, the statistical properties of the DBpedia ontology calculated by Heist et al.~\cite{DBLP:series/ssw/HeistHRP20} were used.

\begin{algorithm}
\caption{Ontology Creation}
\label{alg:ontology-creation}
\begin{algorithmic}
\Procedure{generateClassTree}{numClasses, branchingFactor}
\State $clsURIs \gets $ \Call{generateURIs}{numClasses}
\State $root \gets$ \Call{randomDraw}{clsURIs}
\State $i \gets 0$
\State $workList \gets$ \Call{newList}{ }
\State $result \gets$ \Call{newTree} { }
\State $currentURI \gets root$
\For{$clsURI$ in $clsURIs$}
\If{$clsURI = root$}
\State \Call{continue}{}
\EndIf
\If{$i = branchingFactor$}
\State $currentURI \gets workList.removeFirst()$ 
\State $i \gets 0$
\EndIf
\State $result.addLeaf(currentURI, clsURI)$
\State $i \gets i + 1$
\State $workList.add(clsURI)$
\EndFor
\State \Return $result$
\EndProcedure
\newline
\Procedure{generateProperties}{numProperties, classTree}
\State $properties \gets $ \Call{generateURIs}{numProperties}
\For{$property$ in $properties$}
\State $property.addDomain($ \Call{drawDomainRange}{classTree, 0.25} $)$
\State $property.addRange($ \Call{drawDomainRange}{classTree, 0.25} $)$
\EndFor
\State \Return $properties$
\EndProcedure
\newline
\Procedure{drawDomainRange}{classTree, p}
\State $result \gets classTree.randomClass()$
\While{$Random.nextDouble > p \land \lnot (classTree.getChildren(result) == \emptyset)$}
\State $result \gets randomDraw(classTree.getChildren(result))$
\EndWhile
\EndProcedure
\newline
\Procedure{populateClasses}{numInstances, classTree}
\State $instances \gets $ \Call{generateURIs}{numInstances}
\For{$instance$ in $instances$}
\State $instance.type(classTree.randomClass())$
\EndFor
\State \Return $instances$
\EndProcedure
\end{algorithmic}
\end{algorithm}

\section{Exemplary Analysis}
In order to demonstrate the use of the DLCC benchmark, we compare two flavors of RDF2vec~\cite{DBLP:journals/semweb/RistoskiRNLP19}, two flavors of TransE~\cite{DBLP:conf/nips/BordesUGWY13}, as well as TransR~\cite{DBLP:conf/aaai/LinLSLZ15} and ComplEx~\cite{DBLP:conf/icml/TrouillonWRGB16} embeddings with respect to their capability of separating the classes in the different datasets.

\subsection{Configurations}
For DBpedia, we use version 2021-09. We train RDF2vec in the variants SG and its order-aware counterpart SG$_{oa}$~\cite{DBLP:conf/semweb/PortischP21}.
The embedding files are available via KGvec2go~\cite{DBLP:conf/lrec/PortischHP20}.\footnote{\url{http://data.dws.informatik.uni-mannheim.de/kgvec2go/dbpedia/2021-09/}} 
For the DBpedia embeddings, we used 500 random, duplicate free walks per entity, with a depth of 4, a window of 5, 5 epochs, and a dimension of 200. We used the same parameters for the synthetic gold standard with the exception of $dimension=100$ and $walks=100$ to account for the smaller gold standard size. The embeddings were trained using the jRDF2vec\footnote{\url{https://github.com/dwslab/jRDF2Vec}} framework~\cite{DBLP:conf/semweb/PortischHP20}.

For TransE, we use the variants using the L1 and L2 norm~\cite{DBLP:conf/nips/BordesUGWY13}. TransE, TransR, and ComplEx were trained using the DGL-KE framework\footnote{\url{https://github.com/awslabs/dgl-ke}}~\cite{DBLP:conf/sigir/ZhengSMTYDXZK20}, using the respective default parameters, with 200 dimensions for DBpedia and 100 for the synthetic datasets, as for RDF2vec. The models are publicly available.\footnote{\url{http://data.dws.informatik.uni-mannheim.de/kgvec2go/dbpedia/2021-09/non-rdf2vec/}}

\begin{figure}[t]
    \centering
    \includegraphics[width=\textwidth]{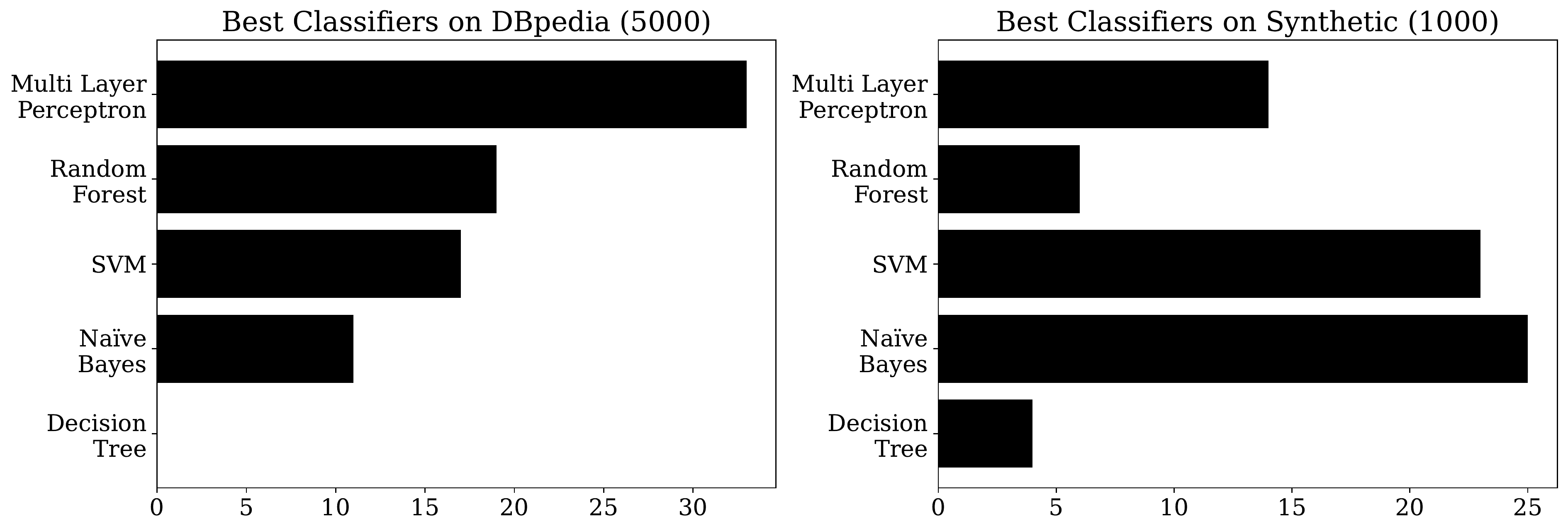}
    \caption{Best Classifiers on the DBpedia and Synthetic Gold Standards. It is important to note that the total number of test cases varies between the two gold standards -- therefore, two separate plots were drawn.}
    \label{fig:best-classifiers}
\end{figure}
\begin{figure}[t]
    \centering
    \includegraphics[width=0.6\textwidth]{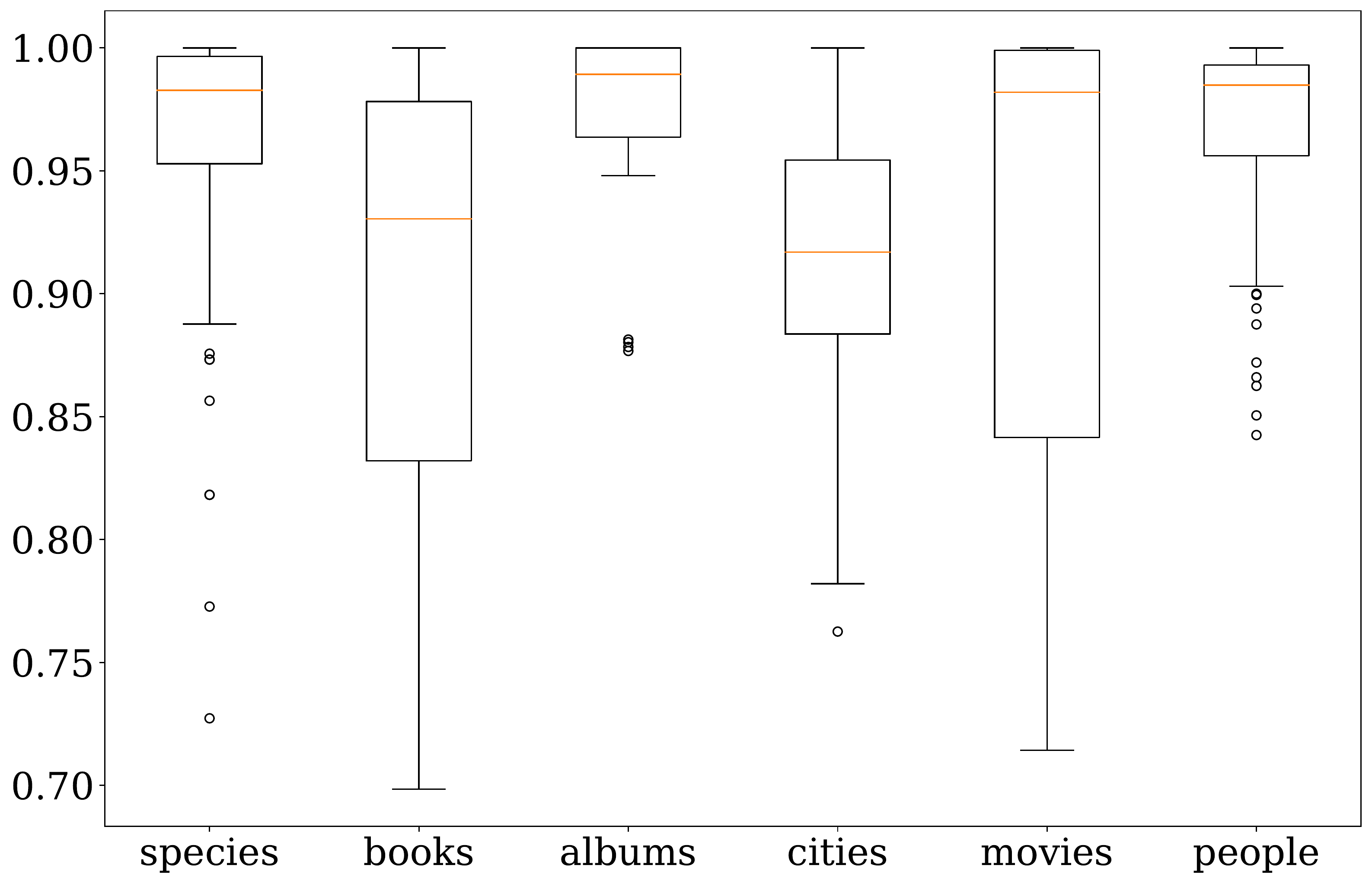}
    \caption{Domain Complexity of the DBpedia Gold Standard (Size Class 5000)}
    \label{fig:domain-complexity}
\end{figure}

\subsection{Results and Interpretation}
The results on the DBpedia gold standard (class size 5,000) and the synthetic gold standard (class size 1,000) are depicted in Tables~\ref{tab:dbpedia-results} and~\ref{tab:synthetic-results}.
For each model and test case, six classifiers were trained (192 classifiers in total). The tables present the results of the best classifiers. 
We performed significance tests (approximated one-sided binomial test) for each test case and approach with $\alpha = 0.05$ to determine whether the accuracy is significantly higher than $0.5$ (random guessing). Since multiple classifiers were trained for each test case, we applied a Bonferroni correction~\cite{DBLP:journals/datamine/Salzberg97} of $\alpha$ to account for the multiple testing problem. On the DBpedia gold standard, all results are significant; on the synthetic gold standard, more insignificant results are observed, particularly for TransR and ComplEx.

Figure~\ref{fig:best-classifiers} shows the aggregated number of the best classifiers for each embedding on each test case. It is visible that on DBpedia, MLPs work best, followed by random forests and SVMs. On the synthetic gold standard, na\"ive Bayes works best most of the time, followed by SVMs and MLPs. The differences can partly be explained by the different size classes of the training sets (MLPs and random forests typically work better on more data).

\begin{table}[t]
\centering
\caption{Results on the DBpedia Gold Standard. The best result for each test case is printed in bold. Listed are the results of the best classifier for each task and model.}
\label{tab:dbpedia-results}
\setlength{\tabcolsep}{4pt}
\footnotesize
\begin{tabular}{llllllllllllllllllll}
\toprule
TC & RDF2vec& RDF2vec$_{oa}$& TransE-L1& TransE-L2& TransR& ComplEx\\\toprule
\\
tc01 & 0.915 & 0.937 & 0.842 & \textbf{0.947} & 0.858 & 0.862\\
tc01 hard & 0.681 & 0.891 & 0.799 & \textbf{0.916} & 0.744 & 0.651\\
tc02 & 0.953 & 0.961 & 0.852 & \textbf{0.970} & 0.832 & 0.853\\
tc02 hard & 0.637 & 0.780 & 0.780 & \textbf{0.849} & 0.693 & 0.608\\
tc03 & 0.949 & \textbf{0.958} & 0.821 & 0.933 & 0.856 & 0.874\\
tc04 & 0.960 & 0.968 & 0.934 & 0.986 & 0.973 & \textbf{0.990}\\
tc04 hard & 0.963 & \textbf{0.984} & 0.814 & 0.912 & 0.855 & 0.935\\
tc05 & 0.986 & \textbf{0.992} & 0.867 & 0.948 & 0.881 & 0.905\\
tc06 & 0.957 & 0.963 & 0.929 & 0.985 & 0.976 & \textbf{0.991}\\
tc06 hard & 0.863 & 0.936 & 0.823 & 0.779 & \textbf{0.964} & 0.933\\
tc07 & 0.938 & 0.955 & 0.930 & \textbf{0.987} & 0.978 & 0.966\\
tc08 & 0.961 & \textbf{0.966} & 0.898 & 0.964 & 0.870 & 0.888\\
tc09 & 0.902 & 0.901 & 0.884 & \textbf{0.938} & 0.879 & 0.883\\
tc09 hard & 0.785 & 0.793 & 0.749 & \textbf{0.848} & 0.758 & 0.776\\
tc10 & 0.947 & 0.958 & 0.957 & \textbf{0.984} & 0.898 & 0.931\\
tc10 hard & 0.740 & 0.737 & \textbf{0.775} & 0.774 & 0.656 & 0.739\\
tc11 & 0.932 & 0.897 & 0.917 & \textbf{0.960} & 0.930 & 0.946\\
tc11 hard & 0.725 & 0.737 & 0.712 & \textbf{0.806} & 0.753 & 0.723\\
tc12 & 0.955 & 0.938 & 0.961 & \textbf{0.984} & 0.879 & 0.894\\
tc12 hard & 0.714 & 0.717 & 0.762 & \textbf{0.765} & 0.659 & 0.710\\\bottomrule
\end{tabular}
\end{table}

\begin{table}[t]
\centering
\caption{Results on the Synthetic Gold Standard. The best result for each test case is printed in bold; statistically insignificant results are printed in italics. Listed are the results of the best classifier for each task and model.}
\label{tab:synthetic-results}
\setlength{\tabcolsep}{8pt}
\footnotesize
\begin{tabular}{lllllllllllllllllll}
\toprule
TC & RDF2vec& RDF2vec$_{oa}$& TransE-L1& TransE-L2& TransR& ComplEx\\\toprule
tc01 & \textbf{0.882} & 0.867 & 0.767 & 0.752 & 0.712 & 0.789\\
tc02 & \textbf{0.742} & 0.737 & 0.677 & 0.677 & \emph{0.531} & \emph{0.549}\\
tc03 & 0.797 & \textbf{0.812} & \emph{0.531} & 0.581 & \emph{0.554} & \emph{0.536}\\
tc04 & \textbf{1.000} & 0.998 & 0.790 & 0.898 & 0.685 & \emph{0.553}\\
tc05 & \textbf{0.892} & 0.819 & 0.691 & 0.774 & 0.631 & 0.726\\
tc06 & 0.978 & 0.963 & 0.898 & 0.978 & 0.888 & \textbf{1.000}\\
tc07 & 0.583 & 0.583 & \emph{0.540} & 0.615 & \textbf{0.673} & \emph{0.518}\\
tc08 & 0.563 & 0.585 & 0.585 & \textbf{0.613} & \emph{0.540} & \emph{0.523}\\
tc09 & 0.610 & \textbf{0.628} & 0.588 & \emph{0.543} & \emph{0.525} & \emph{0.545}\\
tc10 & \textbf{0.638} & 0.623 & 0.588 & 0.573 & \emph{0.518} & \emph{0.510}\\
tc11 & \textbf{0.633} & 0.580 & 0.583 & 0.590 & 0.573 & 0.590\\
tc12 & \textbf{0.644} & 0.614 & 0.618 & \emph{0.550} & \emph{0.513} & \emph{0.540}\\\bottomrule
\end{tabular}
\end{table}

Figure~\ref{fig:domain-complexity} depicts the complexity per domain of the DBpedia gold standard in a box-and-whisker plot. The complexity was determined by using the accuracy of the best classifier of each embedding model without hard test cases (since not every domain has an equal amount of hard test cases). We observe that all domain test cases are similarly hard to solve, whereby the albums, people, and species domain are a bit simpler to solve than the books and cities domain.

In general, we can observe that the results on the DBpedia gold standard are much higher than on the synthetic gold standard. While on the DBpedia gold standard, all but five tasks can be solved with an accuracy above 0.9 (although the cases with hard variants are actually harder than the non-hard ones, and all the five problems with a best accuracy below 0.9 are hard cases), the synthetic gold standard has quite a few tasks (tc07--tc12) which are obviously much harder. For example, it is hardly possible for any of the approaches to learn classes whose definitions involve cardinalities. RDF2vec can produce results slightly above the baseline here because the frequencies of properties appearing in random walks can reflect cardinalities to a certain extent.

Furthermore, we can observe that it seems easier to predict patterns involving outgoing edges than those involving ingoing edges (cf. tc02 vs. tc01, tc08 vs. tc07, tc10 vs. tc09, tc12 vs. tc11), at least for the DBpedia case. Even though the tasks are very related, this can be explained by the learning process, which often emphasizes outgoing directions: In RDF2vec, random walks are performed in forward direction; similarly, TransE is directed in its training process.


For constructors involving a particular entity (tc04 and tc05), we can observe that RDF2vec is clearly better than embedding approaches for link prediction, at least on the synthetic gold dataset. Those tasks refer to \emph{entity relatedness}, for which RDF2vec has been shown to be more adequate~\cite{DBLP:conf/semweb/PortischP21,DBLP:journals/corr/abs-2204-02777}. The picture is more diverse for the other cases.



\subsection{DBpedia Gold Standard vs. Synthetic Gold Standard}
\label{subsec:dbpedia_vs_synthetic}
The results reveal great differences between the gold standards. Many class constructors that are easily learnable on the DBpedia gold standard are hard on the synthetic one. Moreover, the previously reported superiority of RDF2vec$_{oa}$ over standard RDF2vec~\cite{DBLP:journals/semweb/PortischHP22,DBLP:conf/semweb/PortischP21} cannot be observed on the synthetic data.

\begin{figure}[t]
    \centering
    \includegraphics[width=\textwidth]{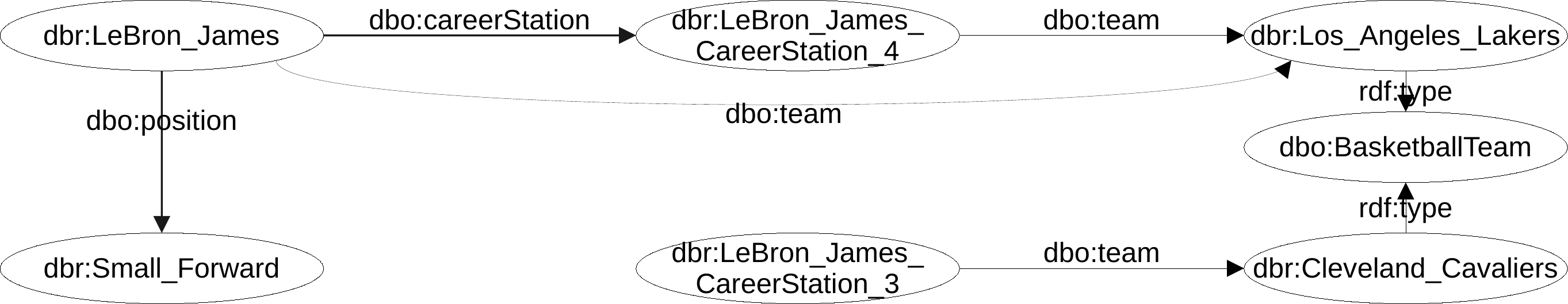}
    \caption{Excerpt of DBpedia}
    \label{fig:excerpt_dbpedia}
\end{figure}

Figure~\ref{fig:excerpt_dbpedia} shows an excerpt of DBpedia, which we will use to illustrate these deviations. The instance \texttt{dbr:LeBron\_James} is a positive example for task tc07 in Table~\ref{tab:sparql-tcs-1}. At the same time, 95.6\% of all entities in DBpedia fulfilling the positive query for positive examples also fall in the class $\exists \texttt{dbo:position}.\top$ (which is a tc01 problem), but only 13.6\% of all entities fulfilling the query for trivial negatives. Hence, on a balanced dataset, this class can be learned with an accuracy of 0.91 by any approach that can learn classes of type tc01. As a comparison to the synthetic dataset shows, the results on the DBpedia test set for tc07 actually overestimate the capability of many embedding approaches to learn classes constructed with a tc07 class constructor.
Such correlations are quite frequent in DBpedia but vastly absent in the synthetic dataset.

The example can also explain the advantage of RDF2vec$_{oa}$ on DBpedia. Unlike standard RDF2vec, this approach would distinguish the appearance of \texttt{dbo:team} as a direct edge of \texttt{dbr:LeBron\_James} as well as an indirect edge connected to \texttt{dbr:LeBron\_James\_CareerStation\_\textit{N}}, where the former denotes the current team, whereas the latter also denotes all previous teams. Those subtle semantic differences of distinctive usages of the same property in various contexts also do not exist in the synthetic gold standard. Hence, the order-aware variant of RDF2vec does not have an advantage here.

\section{Conclusion and Future Work}
In this paper, we presented DLCC, a resource to analyze embedding approaches in terms of which kinds of classes they are able to represent.
DLCC comes with an evaluation framework to easily evaluate embeddings using a reproducible protocol. All DLCC components, i.e., the gold standard, the generation framework, and the evaluation framework, are publicly available.\footnote{Dataset DOI: \texttt{10.5281/zenodo.6509715}} 

We have shown that many patterns using DL class constructors on DBpedia are actually learned by recognizing patterns with other constructors correlating with the pattern to be learned, thus yielding misleading results. This effect is less prominent in the synthetic gold standard. We showed that certain DL constructors, such as cardinality constraints, are particularly hard to learn.

In the future, we plan to extend the systematic evaluation to more embedding approaches, including the flavors of RDF2vec, which were published more recently~\cite{DBLP:conf/semweb/PortischP21,DBLP:journals/corr/abs-2204-02777,DBLP:conf/dexaw/SteenwinckelVBW21}. The synthetic dataset generator also allows for more interesting experiments: We can systematically analyze the scalability of existing approaches or study how variations in the synthetic gold standard (e.g., larger and smaller ontologies) influence the outcome.

%
%

\bibliographystyle{splncs04}
\bibliography{references}
\end{document}